\documentclass[]{spie}  

\usepackage{amsmath,amsfonts,amssymb}
\usepackage{graphicx}
\usepackage[colorlinks=true, allcolors=blue]{hyperref}
\usepackage{lineno}

\usepackage{amsfonts}
\usepackage{amsmath}

\usepackage{array}
\usepackage{multirow}

\title{Achieving state-of-the-art performance in the Medical Out-of-Distribution (MOOD) challenge using plausible synthetic anomalies}

\author[a]{Sergio Naval Marimont}
\author[a,b]{Giacomo Tarroni}
\affil[a]{CitAI Research Centre, City, University of London, London, UK}
\affil[b]{BioMedIA, Imperial College, London, UK}

\authorinfo{Further author information: (Send correspondence to S.N.M)\\S.N.M.: E-mail: sergio.naval-marimont@city.ac.uk\\  G.T.: E-mail: giacomo.tarroni@city.ac.uk}

\begin{document} 
\maketitle


\begin{abstract}
\textbf{Purpose:} The detection and localization of anomalies is one important medical image analysis task. Most commonly, Computer Vision anomaly detection approaches rely on manual annotations that are both time consuming and expensive to obtain. Unsupervised anomaly detection, or Out-of-Distribution detection, aims at identifying anomalous samples relying only on unannotated samples considered normal. In this study we present a new unsupervised anomaly detection method. 

\noindent
\textbf{Approach:} Our method builds upon the self-supervised strategy consisting on training a segmentation network to identify local synthetic anomalies. Our contributions improve the synthetic anomaly generation process, making synthetic anomalies more heterogeneous and challenging by 1) using complex random shapes and 2) smoothing the edges of synthetic anomalies so networks cannot rely on the high gradient between image and synthetic anomalies. In our implementation we adopted standard practices in 3D medical image segmentation, including 3D U-Net architecture, patch-wise training and model ensembling.

\noindent
\textbf{Results}: Our method was evaluated using a validation set with different types of synthetic anomalies. Our experiments show that our method improved substantially the baseline method performance. Additionally, we evaluated our method by participating in the Medical Out-of-Distribution (MOOD) Challenge held at MICCAI in 2022 and achieved first position in both sample-wise and pixel-wise tasks.

\noindent
\textbf{Conclusions:} Our experiments and results in the latest MOOD challenge show that our simple yet effective approach can substantially improve the performance of Out-of-Distribution detection techniques which rely on synthetic anomalies.

\end{abstract}
\section{Introduction}
Supervised deep learning approaches achieve state of the art performance in many medical image analysis tasks and are being progressively incorporated in the clinical practice \cite{Litjens2017}. However, supervised deep learning methods have important limitations. First, they rely on large and diverse annotated datasets which are difficult and expensive to obtain. Ownership and patient privacy create important constraints to the availability of medical datasets. Furthermore, deep learning requires images to be manually annotated by clinical experts in a process that is both onerous and costly \cite{Litjens2017}. 

One important task in medical image analysis is the detection and localization of anomalies, which we explore in this work. In this context, anomalies comprise not expected anatomies, that can be pathological, or image quality issues created during the image acquisition or reconstruction processes. In anomaly detection, supervised learning methods are limited to the anomalies annotated and cannot be expected to generalize to unseen/unannotated anomalies. Similarly, supervised methods are known to generate unreliable predictions for Out-of-Distribution (OoD) samples. OoD refers to samples that are from a different probability distribution from the one observed during training. For example, sufficiently different acquisition protocols or different populations (not present in the study cohort) could trigger unreliable predictions by supervised methods \cite{Hendrycks2018}.  

OoD or unsupervised medical anomaly detection methods address some of these challenges by seeking to identify anomalous images leveraging only a training set of images of normal subjects \cite{MOOD2020}. First, they do not require annotations, so they can directly be implemented using datasets from known healthy subjects. Second, unsupervised methods \textit{learn} the expected image quality and/or normal subject anatomies, and identify as OoD samples those that do not conform with the learnt distribution. Consequently they can be expected to identify two relevant scenarios: 1- pathological anatomies and 2- images that, without pathologies, just differ from the distribution used during training, where predictions from supervised methods can be unreliable.

One of the most promising OoD detection strategies consists in corrupting images obtained from normal/healthy subjects with localized synthetic anomalies and then training deep segmentation networks to localize these synthetic anomalies \cite{Tan2020, LiC2021}. One key aspect to these methods is the design of the synthetic anomaly generation process which needs to render naturally-looking and challenging anomalies. For example, by replacing an image section with a patch obtained from a different image (foreign patch) it is possible to obtain naturally looking-images with anomalous anatomies. Approaches based on this strategy \cite{Tan2020,Cho2021} won the 2020 and 2021 editions of the Medical Out-of-Distribution Challenge (MOOD)\footnote{\href{http://medicalood.dkfz.de/web/}{\texttt{http://medicalood.dkfz.de/web/}}} held at the Medical Image Computing and Computer Assisted Interventions (MICCAI) conference \cite{MOOD2020}. The MOOD Challenge evaluates OoD methods in two medical modalities, abdominal CT scans and brain MRIs. Participants are only provided with images considered normal and methods are compared in their ability to identify and localize a broad set of anomalies including real pathologies, image quality issues and synthetic anomalies. To avoid dishonest methods that rely on test set annotations, the challenge organizers keep the test set confidential at all times.

In this work, we build upon the existing synthetic anomaly OoD detection literature and propose a set of simple yet highly effective improvements to the  anomaly generation processes. First, we introduce random anomaly shapes (instead of the usually-adopted cuboids) so that synthetic anomalies are more diverse. Second, we propose to smooth edges between the foreign patch and image so networks cannot simply rely on the high gradients between image and foreign patch to identify anomalies. Finally, we introduced common image segmentation practices and architectures (e.g. 3D UNets, self-supervision, ensembling, sliding window inference,...) to improve upon baseline methods. We validated our contributions both in a synthetic anomaly dataset and by participating in the MOOD 2022 challenge, where we obtained the first position in both sample-wise and pixel-wise rankings.

\subsection{Related Works:} 

Out-of-Distribution detection aims at identifying anomalies 
 by leveraging only samples from a distribution of healthy anatomies. Self-supervised learning approaches fit naturally the task because they are designed to model the data distribution without requiring explicit annotations. 

Within self-supervised learning approaches, generative models have been broadly studied in the literature. Generically, generative models are trained to learn the anomaly-free distribution and anomalies are identified measuring the distance between a test sample and the learnt healthy distribution. One class of generative models is Variational Auto-Encoders (VAE) \cite{Kingma2013}. VAEs learn to encode images in a latent space and to reconstruct images from the generated representations. Vanilla OoD detection approaches with VAEs assume that models of the healthy-only distribution will not be able to reconstruct anomalous samples, and consequently pixel-wise residuals between original image and reconstruction can be used as anomaly score (AS) \cite{Baur2020}. In Zimmerer et al. \cite{Zimmerer2019}, authors found that the KL-divergence term in the VAE loss function could be used to detect anomalous samples. Furthermore the authors proposed to localise anomalies using gradients of the KL-divergence term w.r.t. pixels. Chen et al. \cite{Chen2020} suggested using the gradients of the VAE loss function w.r.t. pixels values to iteratively \textit{heal} the images and turn them into in-distribution samples, in a so-called \textit{restoration} process. Generative Adversarial Networks (GANs) \cite{Goodfellow2014} is another notorious class of generative models. OoD detection with GANs also assumes that anomalous samples cannot be sampled from a GAN model trained from the anomaly-free distribution. With this strategy, f-AnoGAN \cite{Schlegl2019} proposes to find representations maximally consistent with the test image and it identifies as anomalies the areas where the test image differs from the GAN samples. 2-stage generative models have also been proposed for OoD detection \cite{Wang2020, Naval2020, PinayaW2022a, PinayaW2022d, Naval2023}. In a 2-stage generative model, the first stage encodes images into spatial latent representations, typically using Vector Quantized-VAE \cite{Oord2017}. In the second stage, a generative model is used to learn the likelihood of latent representations. Auto-Regressive modelling \cite{Oord2017}, Diffusion models \cite{RombachR2022} or Masked Image Modelling \cite{ChangH2022} are different choices for the second stage. OoD detection methods with 2-stage generative models are based on the assumption that the second stage model will predict low-likelihood in anomalous samples. Leveraging the generative capabilities of the second stage, latent variables with low-likelihood can be replaced with in-distribution samples to generate \textit{healed} reconstructions. Anomaly localization is then achieved comparing original images and \textit{healed} reconstructions. One drawback of generative models for localization is that they define anomaly scores comparing image intensities of test samples and reconstructions and consequently the localization performance decreases in those anomalies with intensities similar to healthy anatomies.

A different self-supervised strategy for OoD detection is to identify synthetic anomalies as a pretext task. The idea is turning OoD detection into a fully-supervised segmentation task to identify local corruptions. If synthetic anomalies present sufficient variety and complexity, the trained network can be expected to generalize to unseen anomalies. Foreign Patch Interpolation (FPI) \cite{Tan2020} proposes to generate local corruptions into a healthy image copying foreign patches obtained from a second healthy image, similarly to what was proposed in CutMix \cite{Yun2019} as an augmentation method. By leveraging the texture obtained from in-distribution images, the generated anomalies exhibit a natural aspect. Furthermore, instead of locally replacing all pixels, the corrupted image is obtained using a linear interpolation between original and foreign patch with an interpolation factor $\alpha$, practically defining the weight of the foreign patch. FPI uses a 2D segmentation network trained to estimate the interpolation factor $\alpha$ as a measure of how anomalous are specific pixels. With this strategy FPI achieved the first position in the MOOD Challenge 2020 \cite{MOOD2020}. Although FPI anomalies have natural looking textures, the edges between original image and foreign patch can easily reveal the corruption location and networks can learn to identify FPI patches relying on high gradients present on those edges \cite{Tan2021}. While this supports the identification of many synthetic anomalies, the trained methods might not be able to identify medical anomalies with softer edges. Poisson Image Interpolation (PII) \cite{Tan2021} was proposed to address this limitation of FPI. In PPI, foreign patches are blended with the training images using Poisson image editing \cite{Perez2003} to achieve more natural looking synthetic anomalies. Natural Synthetic Anomalies (NSA) \cite{Schluter2022} builds upon PII to add constraints on the anomaly location, and foreign patches are further augmented. nn-OOD \cite{Baugh2022} adapts the successful image segmentation methodology described in nn-UNet \cite{Isensee2021} and comparatively evaluates previously described methods to generate synthetic anomalies. Cho et al. \cite{Cho2021} won the 2021 MOOD challenge also using the self-supervised strategy of identifying synthetic anomalies. Their approach introduces additional augmentations to the foreign patches as proposed in CutPaste \cite{LiC2021} and features a 3D U-Net architecture \cite{Ronneberger2015} instead of the 2D one used in FPI/PII/NSA approaches.

\subsection{Contributions:} 
We designed and implemented an unsupervised anomaly detection method for medical images following the synthetic anomaly segmentation strategy, building on methods like FPI and PPI. Our technique was tested in the most important benchmark for OoD detection in medical images, the MOOD Challenge held at MICCAI where we secured first place in both sample- and pixel-wise tasks. The main contributions of our paper are the following:
\begin{itemize}
    \item We propose a series of modifications to the synthetic anomaly generation process aiming to prevent segmentation networks from learning spurious patterns that limit their capacity to generalize to unseen anomaly types. Our contributions include the introduction of random shapes and smoother blending of synthetic anomalies into the training images.
    \item We systematically report the impact of these modifications on a dataset of synthetic anomalies (different from the ones used for training), using a similar validation strategy to that of FPI \cite{Tan2020}.
    \item We make public the implementation of the architecture and pipeline we adopted for our submission to the MOOD 2022.
\end{itemize}

Section \ref{Synthetic anomaly generation process} presents the methods proposed to generate synthetic anomalies. In subsections \ref{Network architecture}, \ref{Training objective}, \ref{Anomaly Scores} we describe respectively network architecture, training procedure and the definition of the Anomaly Score used to identify anomalies. In section \ref{Experiments and Results} we discuss the experiments used to validate our method and the results obtained. Finally, in section \ref{Conclusions} we present our conclusions.

\section{Methods}\label{methods}
The method proposed aims at identifying anomalous images leveraging only a dataset containing normal images. To this end, we introduce synthetic anomalies to the normal images and train a segmentation network to identify them. The process to generate rich, diverse and realistic synthetic anomalies is key for the network to generalize to unseen types of anomalies. Consistently with previous literature\cite{Tan2020, Tan2021}, we generate synthetic anomalies leveraging textures extracted from the training dataset. To this end, we extract image patches (named \textit{foreign patches}) from the healthy images and paste them into random locations through interpolation between original intensity values and foreign patch intensities.

\subsection{Synthetic anomaly generation process}\label{Synthetic anomaly generation process}
\begin{figure}
\centering
\includegraphics[width=0.9\textwidth]{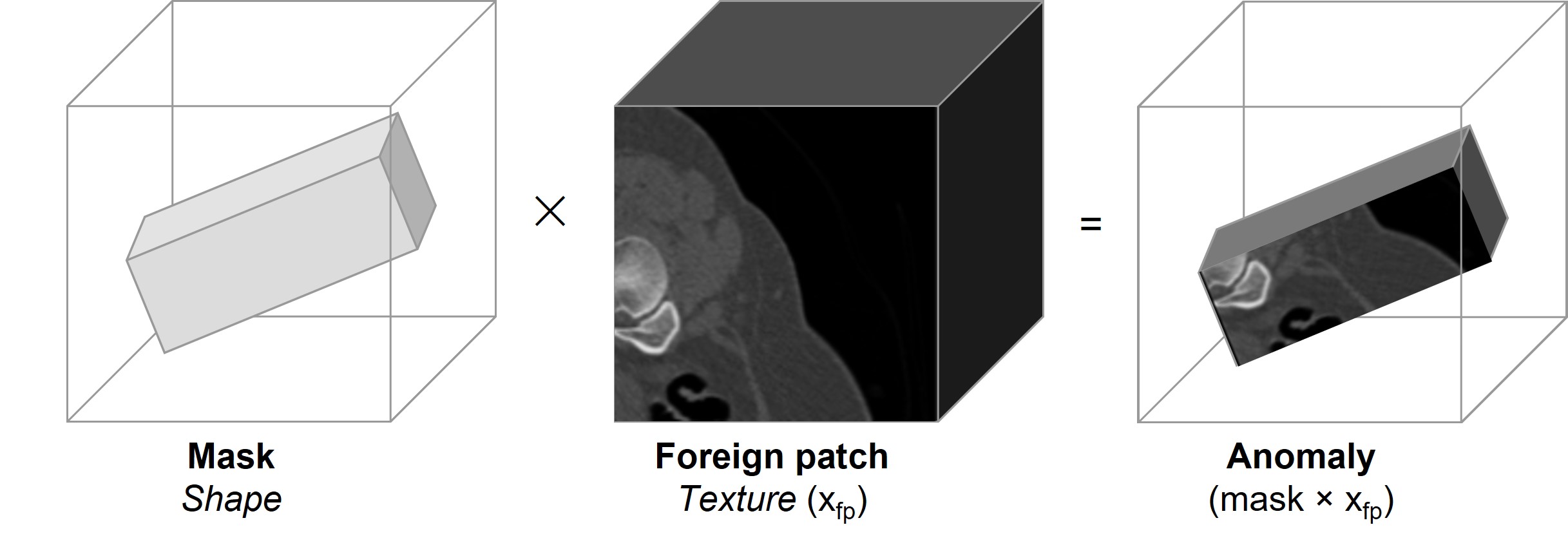}
\caption{Texture (foreign patch) and shape components (mask) of the synthetic anomaly generation process.} \label{fig1}
\end{figure}

We describe the anomaly generation process (shown in Figure \ref{fig1}) from three characteristics: anomaly texture, shape and interpolation mechanism. The implementation of our synthetic anomaly generation process is publicly available\footnote{\href{https://github.com/snavalm/mood22}{\texttt{https://github.com/snavalm/mood22}}}.

\textbf{Anomaly texture:} With the objective of obtaining rich and heterogeneous textures for our synthetic anomalies, we leverage textures obtained from the training dataset itself. To this end, we randomly sample 3D patches $x_{fp}$ with dimensions $( W_{fp}, H_{fp}, D_{fp} )$ and add them to a memory bank containing $K$ 3D image patches. During training we will be using one randomly sampled patch from the memory bank, named foreign patch, to generate anomalies in each image processed. In order to avoid network overfitting to a fixed set of patches we replace patches in memory bank sampling one new patch from each training image processed.

Similarly to the approach used by Cho et al.\cite{Cho2021}, before using a foreign patch to generate synthetic anomalies we augment it with random Gaussian noise, random intensity shifts and random rotations. Augmentations aim at increasing the diversity of the generated anomalies.

\begin{figure}
\centering
\includegraphics[width=0.7\textwidth]{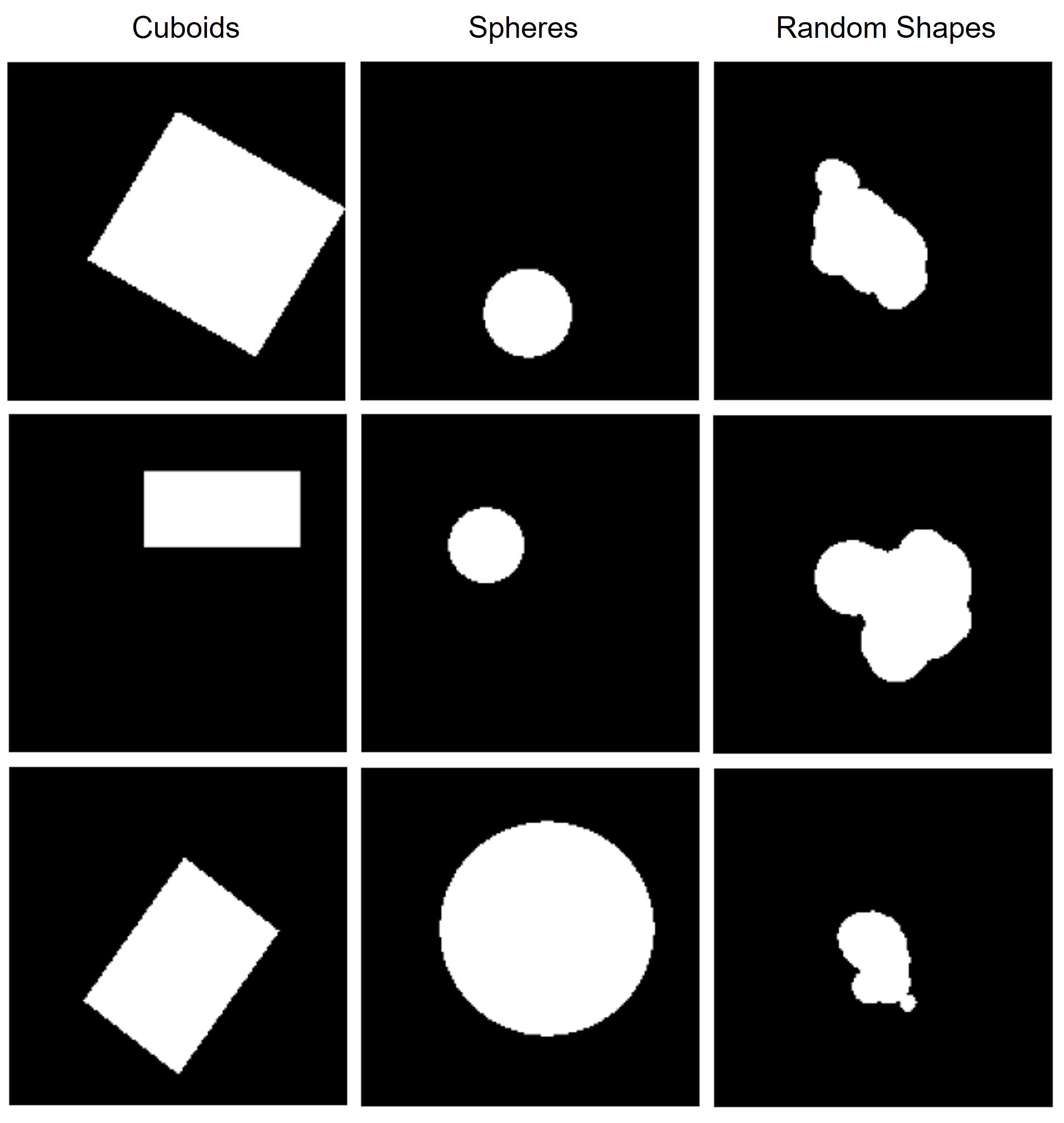}
\caption{Examples of augmented shapes for each category used in the anomaly generation process. Note that these are axial 2D slices of the actual 3D shapes used in our method.} \label{fig2}
\end{figure}

\textbf{Anomaly shape:} The second characteristic of synthetic anomalies is the shape, referred to as \textit{mask} in Figure \ref{fig1}. While previous approaches described in the literature use square patches, we hypothesized that adding diversity and complexity to the shapes of synthetic anomalies could enhance the anomaly detection method. In this work we explored using different shapes, which include rectangular cuboids, spheres and random 3D shapes. In all our experiments random rectangular cuboids were augmented with random rotations. Examples of the different shapes used in our approach can be found in Figure \ref{fig2}.

We generate random 3D shapes using a procedure that simulates a brush moving on a random walk along a 3D \textit{canvas}. Specifically, the simulated random brush is characterized by a position in the 3D coordinate space $(x,y,z)$ and a spherical brush radius $r$. The 3D \textit{canvas} has the same dimensions as the foreign patch, i.e. $( W_{fp}, H_{fp}, D_{fp} )$ and it is initialized with $0$ values. The brush starts in the centre of the 3D \textit{canvas}, i.e. $( W_{fp} / 2 , H_{fp} / 2, D_{fp} / 2 )$ and randomly moves for $S$ steps. At each step, the voxels falling within the spherical brush are set to $1$. The brush position and radius follow a random walk with Gaussian steps. Specifically, at each step we sample $(\Delta x, \Delta y, \Delta z)$ and $\Delta r$ from a normal distribution $\mathcal{N}(0,\sigma)$ and set the new positions and radius applying the sampled increments. We followed the above procedure with multiple hyper-parameters of number of steps $S$, initial radius $r$ and  variance $\sigma$ to generate 300 random shapes that we utilized during training. To increase their diversity, the random shapes are further augmented with random affine transformations.  

\textbf{Interpolation mechanism:} Prior to interpolation, we randomly select a location for the synthetic anomaly. It is important to note that foreign patches, with size $( W_{fp}, H_{fp}, D_{fp} )$, are designed to be smaller than training images, with size $( W, H, D )$, i.e.  $W_{fp} < W, H_{fp} < H, D_{fp} < D$. Our objective is two-fold: first, to ensure the creation of local corruptions instead of global corruptions. For example, if the foreign patch  spans most of the image, the definition of anomaly would be ambiguous as the original image contour could be defined as the anomaly. Second, small 3D foreign patches reduce computations required to generate the anomalies, alleviating the cost of the multiple affine transformations for shape and texture components. The location of the foreign patch is uniformly sampled so the foreign patch fits within the training sample. 

Once a location is determined, training samples (i.e. images with synthetic anomalies) $x'$ are created by replacing original pixel intensities $x$ with a linear interpolation between original intensities $x$ and foreign patch intensities $x_{fp}$ described in the equation below \cite{Tan2020}.

$$
x' = x \times (1 - \alpha) + x_{fp} \times \alpha
$$ 

The $\alpha$ factor that controls the weighting of the foreign patch intensities $x_{fp}$ is sampled uniformly from the range $[ 0.3,1]$. Diverse $\alpha$ factors allow to generate a range of anomalies, from obvious to more subtle, which should mimic the nature of naturally occurring anomalies. 

Additionally, as proposed in the FPI method \cite{Tan2020}, anomalies in brain MRI experiments are generated only in foreground areas. In our experiments the foreground was identified as the voxels with intensities greater than $0$. 

\textbf{Anomaly edge smoothing:} The anomaly generation process proposed described so far in previous sections creates high gradients at the edges of the synthetic anomalies, i.e. where there is a transition between original to corrupted voxels. We hypothesised that networks can learn to leverage these high gradients to identify the synthetic anomalies. By relying on the spurious correlation between high gradients and anomalies our method would fail to identify naturally occurring anomalies, or complex synthetic anomalies, that are not necessarily delimited by high gradient boundaries. 

\begin{figure}[!h]
\centering
\includegraphics[width=0.8\textwidth]{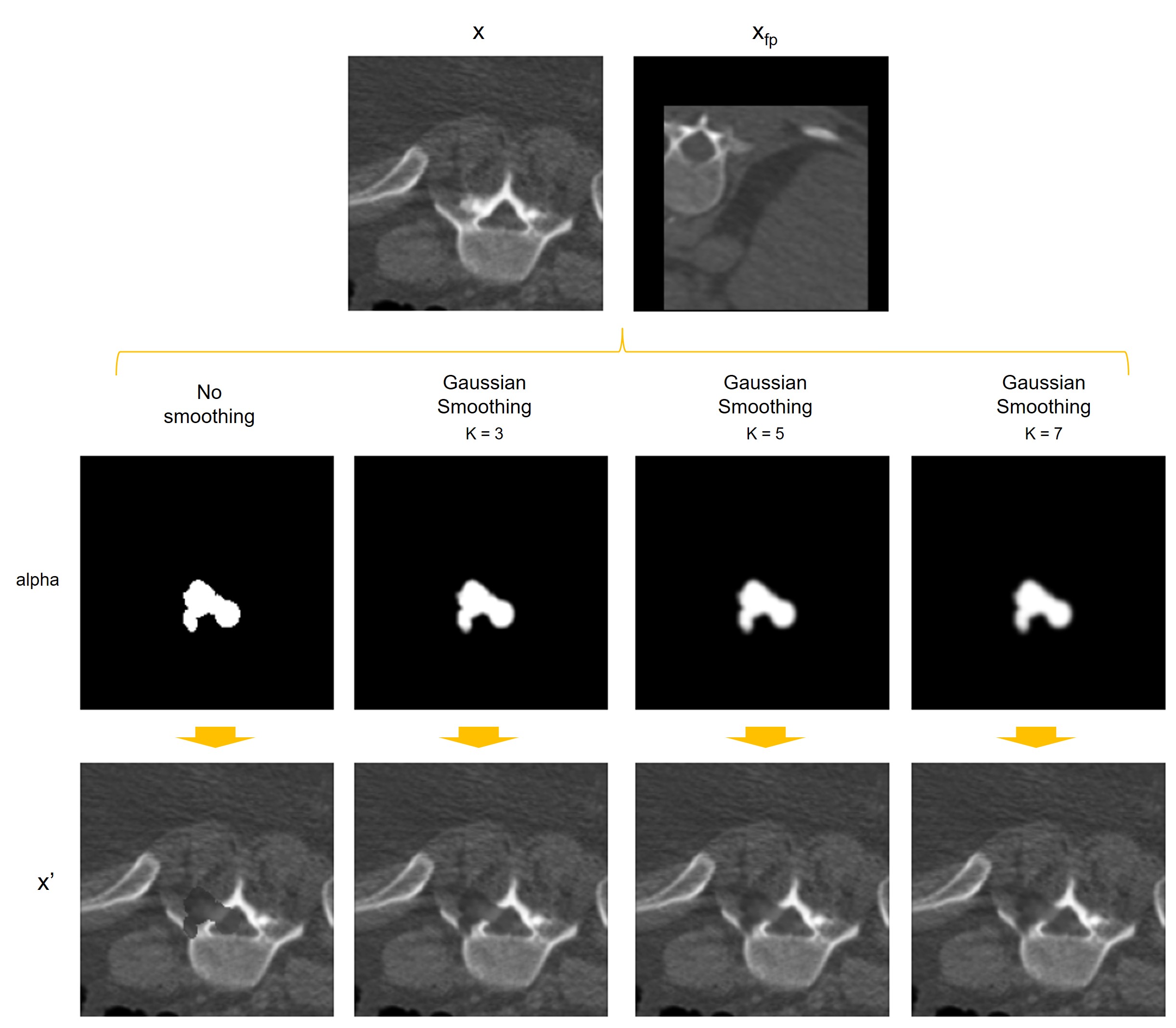}
\caption{Qualitative comparisons of different degrees of anomaly edge smoothing. The original image is $x$, the foreign patch $x_{fp}$, the smoothed shape component $\alpha_s$ and the locally corrupted image $x'$, respectively.}\label{fig3}
\end{figure}

We introduce a simple yet effective strategy to alleviate this issue. Our strategy consists on smoothing the edges of the shape prior to the interpolation so the interpolation factor gradually increases from $\alpha_s = 0$ (i.e. no anomaly) to $\alpha_s = \alpha$ in the centre of the anomaly. We propose to use Gaussian filters to smooth the shape component. The size of the Gaussian kernel $K$ is selected randomly during the anomaly generation process among size values $3,5,7$. Figure \ref{fig3} illustrates the anomaly edge smoothing results compared to the original foreign patch interpolation process.

\subsection{Network architecture} \label{Network architecture}

\begin{figure}
\centering
\includegraphics[width=1\textwidth]{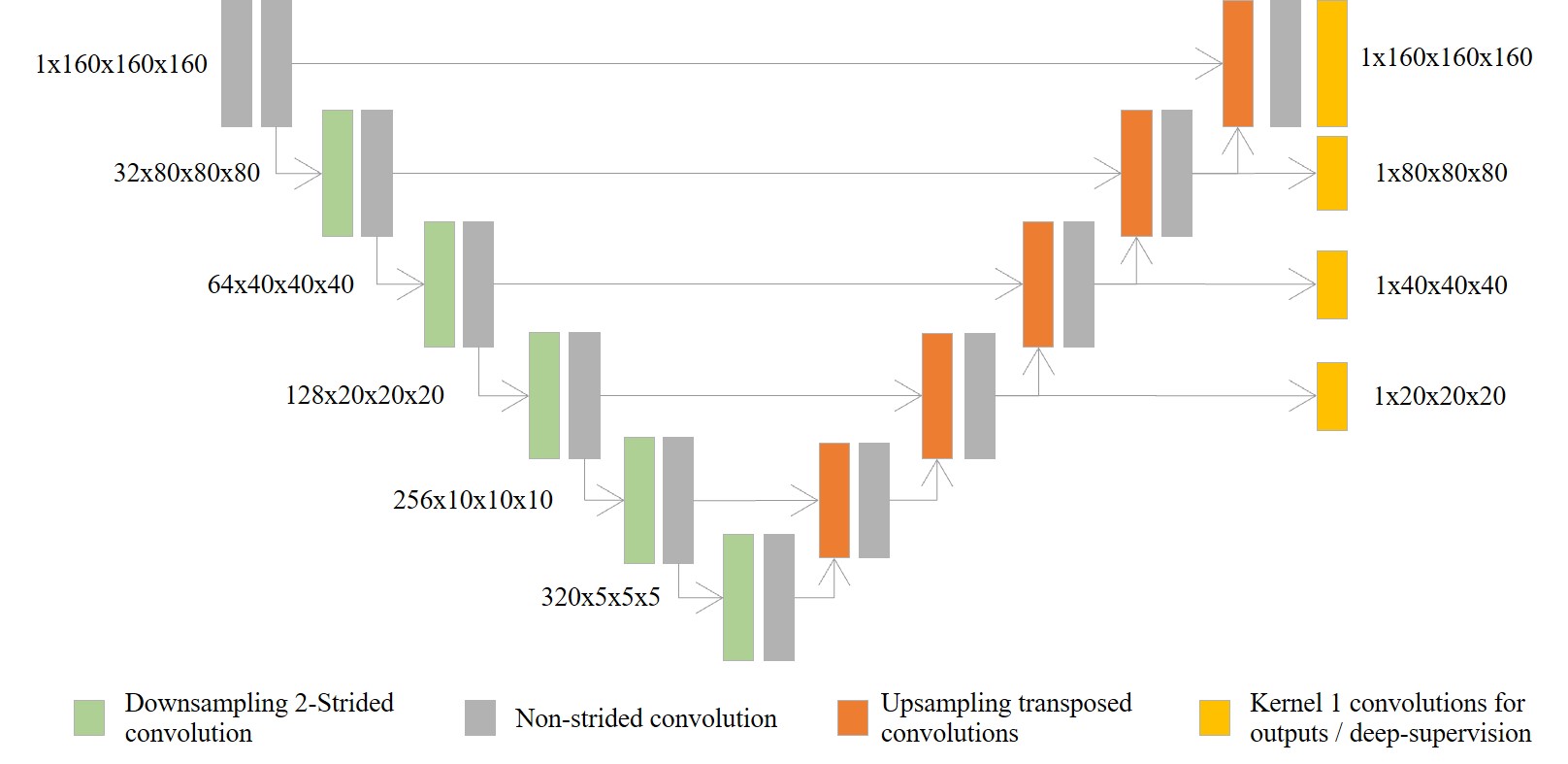}
\caption{The U-Net segmentation network architecture adopted in our implementation.} \label{fig4}
\end{figure}

We used a 3D U-Net architecture \cite{Ronneberger2015} in our experiments. The U-Net architecture is symmetric, with 5 down-sampling and up-sampling blocks each consisting of 2 groups of convolution, batch normalization and non-linearity. We use Leaky ReLU as non-linearity and sigmoid as final activation. Downsamplings and upsamplings are implemented with 2-strided convolutions and transposed convolutions respectively. Additionally we use deep supervision in the outputs of the last 4 blocks. The network architecture is detailed in the diagram in Figure \ref{fig4}. 

\subsection{Training objective}  \label{Training objective}
As proposed in \cite{Tan2020}, we found in our experiments that training the network to estimate the interpolation factor $\alpha$ improves the evaluation metrics. Specifically, we used binary cross-entropy loss pixel-wise: 

$$
\mathcal{L}_{bce} = -\alpha \times \log \big( f(x')\big) -  (1 - \alpha) \times \log \big(1 - f(x')\big) 
$$ 

where $\alpha_s$ is the interpolation factor previously introduced and $f(x')$ is the pixel-wise output of the network for the locally corrupted image $x'$.

\subsection{Anomaly scores}  \label{Anomaly Scores}

At test time, for unsupervised anomaly localization tasks, we use directly the output of our network trained to identify synthetic anomalies. Consequently, the pixel-wise anomaly score (AS) is $ AS_{pixel} = f(x) $.

For image classification tasks (i.e. separating normal and anomalous samples at test time), we define a sample-wise AS relying on the pixel-wise predictions. Specifically $AS_{sample} = \sum^{\mathbf{M}}_i \big(f(x_i)\big) / 100$, where $\mathbf{M}$ is the subset of 100 pixels with higher pixel-wise anomaly scores in the image.

\section{Experiments and Results} \label{Experiments and Results}
\subsection{Datasets:} We evaluated our approach participating in the Medical Out-of Distribution (MOOD) Challenge 2022 \cite{MOOD2020}. The challenge consists of sample- and pixel-wise classification tasks using two training datasets:
\begin{itemize}
    \item \textbf{Brain}: 800 MRI scans obtained from the HCP \cite{VanEssen2012} dataset. The dataset contains 3T MR imaging data from a cohort of healthy young participants (22-35 years old) who were scanned using the same equipment and protocol. Scans with spatial dimensions $(256 \times 256 \times 256)$ were provided by the challenge organizers.
    \item \textbf{Abdominal}: 550 Abdominal CT scans obtained from a colonography study \cite{Johnson2008}. The study, which was performed in 15 centers, included male and female participants aged 50 years old or older scheduled for a colonoscopy screening and which had not had a colonoscopy in the previous 5 years. Images were acquired with standard bowel preparation, stool and fluid tagging, mechanical insufflation and multi-detector row CT scanners (16 or more rows). Images provided in the training set might included polyps, however these were not considered abnormal. Abdominal CT scans with spatial dimensions $(512 \times 512 \times 512)$ were provided by the challenge organizers.
\end{itemize}

Both training datasets were visually checked and pre-processed by the MOOD challenge organizers, who provided the scans with intensities normalized in the range $[0,1]$. We noted that abdominal images had diverse spacing so we resampled the relative dataset to a common isotropic spacing of 1 mm. 

The MOOD challenge test set contains anomaly-free images, images with natural anomalies and images with synthetic anomalies. It is important to note that the MOOD challenge test set is kept confidential at all times to prevent participants from overfitting their methods to the types of anomalies present in the test set. Similarly, the challenge organizers only provide feedback on the aggregated quantitative performance for the final submission of each participant, which prevents the evaluation of different configurations of a given method. Furthermore, the MOOD image pre-processing pipeline is also kept confidential to prevent participants to leverage other datasets when training or evaluating their models.

\begin{figure}
\centering
\includegraphics[width=0.9\textwidth]{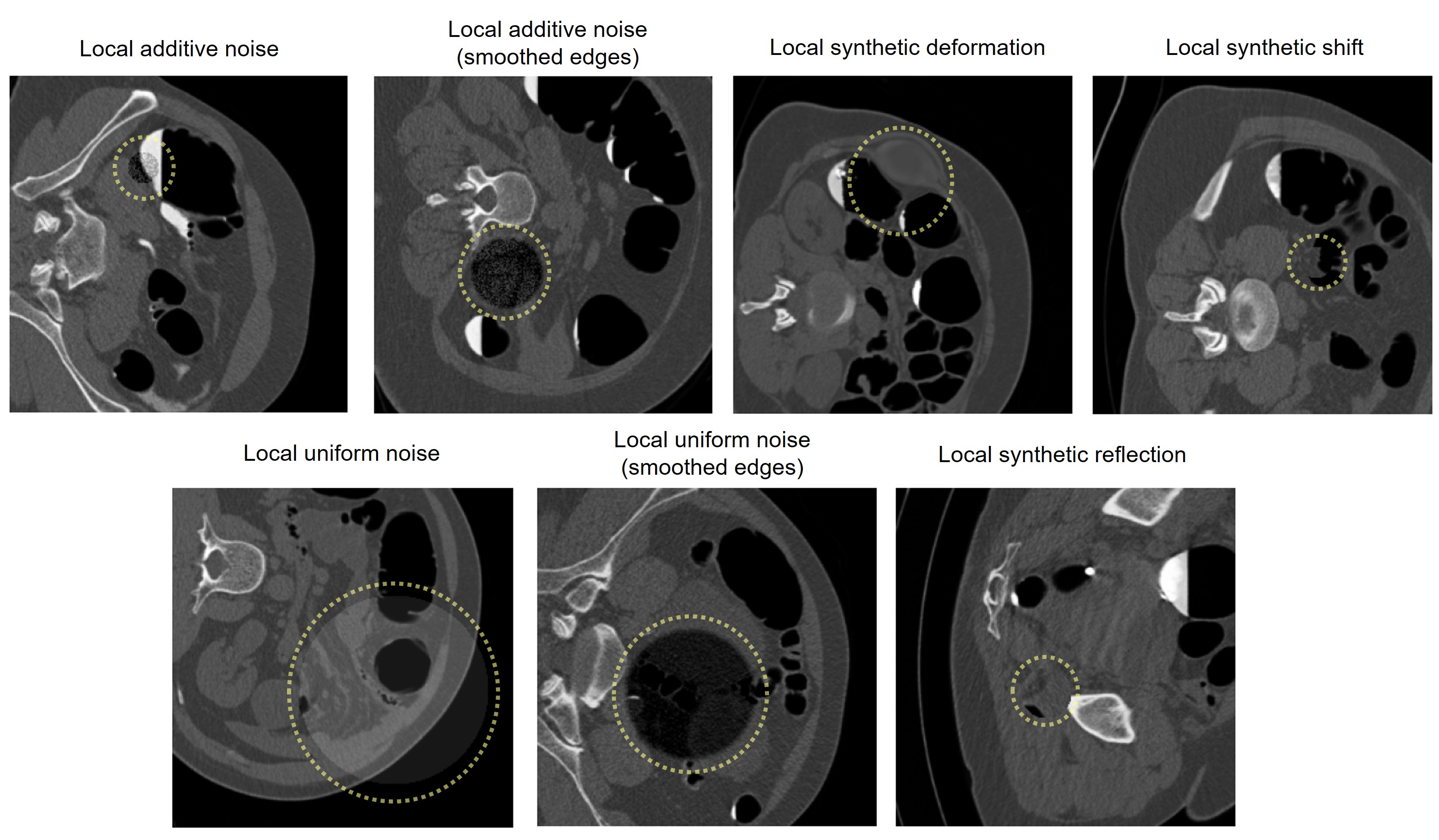}
\caption{Examples for each type of synthetic anomaly (localised inside the yellow dotted line) used in the validation set.} \label{fig5}
\end{figure}

Given the MOOD constraints, we decided to develop a validation dataset leveraging images provided by the MOOD organizers. Specifically, we set aside $1/5$ of each of the two provided datasets for evaluation. Using these evaluation images we built a validation set with synthetically-generated anomalies. 
To this end, we used the same types of validation anomalies proposed in the paper by Tan et al. \cite{Tan2020}, for which the implementation is openly available \footnote{\href{https://github.com/jemtan/FPI/blob/master/synthetic/example_synthesizing_outliers_mood.ipynb}{\texttt{https://github.com/jemtan/FPI/blob/master/synthetic/example\_synthesizing\_outliers\_mood.ipynb}}}. Examples of each of the synthetic anomaly types introduced in the validation set are included in Figure \ref{fig5}.

Additionally, we introduced new types of validation anomalies, i.e. local additive noise (smoothed edges) and local uniform noise (smoothed edges), where the edge of the anomaly was smoothed using the methodology introduced in subsection \ref{Synthetic anomaly generation process}. The rationale for including these new types of anomalies in the validation set was to confirm that methods trained with non-smoothed edges were not generalizing to synthetic anomalies with smooth edges. The composition of each of the two validation sets (brain MRI and abdominal CT) comprised is reported in Table \ref{table:validation_comp}.

\begin{table}[h!]
\centering
\caption{Validation dataset samples by anomaly types}\label{Table:tab0}
\begin{tabular}{p{7cm} >{\centering\arraybackslash}p{2cm} }
\hline
  &  N  \\
\hline
Healthy / uncorrupted images & 50  \\
Local additive noise & 30  \\
Local additive noise (smoothed edges) & 30  \\
Local synthetic deformations & 30  \\
Local synthetic reflections & 30  \\
Local synthetic shifts & 30  \\
Local uniform noise & 30  \\
Local uniform noise (smoothed edges) & 30  \\
\hline
Total & 260 \\
\end{tabular}
\label{table:validation_comp}
\end{table}

\subsubsection{Experimental set-up:}

In all our experiments we trained our network using randomly sampled image patches of size $(160 \times 160 \times 160)$ for 35.000 training steps. We used AdamW \cite{Loshchilov2017} optimizer with OneCycle learning rate scheduler \cite{Smith2019} with a maximum learning rate of $10^{-3}$ and weight decay of $10^{-5}$. We used Pytorch \cite{pytorch} and MONAI \cite{monai} libraries. Our implementation is made publicly available\footnote{\href{https://github.com/snavalm/mood22}{\texttt{https://github.com/snavalm/mood22}}}. 

At inference time, we use sliding window-based inference with 0.5 patch overlap and Gaussian weighting of patches. In our final submission to the MOOD 2022 challenge we ensembled 10 models trained with 5 data folds. In each data fold we trained two models, one without mask edge smoothing and one with mask edge smoothing.

\subsubsection{Performance evaluation:} 
To assess the impact of the proposed contributions we run experiments using the abdominal dataset. We focused on the pixel-wise classification task (i.e. segmentation of the anomalies) and used average precision (AP) as evaluation metric as in the MOOD challenge evaluation process. 

Our baseline model was trained with random cuboid shapes without edge smoothing on anomalies and anomalies with size  $W_{fp} = H_{fp} = D_{fp} = 64$. We evaluated the trained model first on the validation set excluding smoothed edge anomalies (total N=200). Then we introduced the additional 60 samples with smoothed edges. The AP decreased from $0.845$ to $0.740$. This suggests that the model relies on the high gradients in edges to identify anomalies. Furthermore, we trained a new model with both hard and smoothed edges and it achieved AP of $0.681$. These results (together with additional details) are presented in Table \ref{tab:first ablation}. Through qualitative analysis we found that the new model performed comparatively worse in hard edge anomalies but improved in the more challenging anomalies in the validation set.

\begin{table}[!h]
\centering
\caption{Ablation analyses on including smoothed anomalies in the abdominal validation set. Metric is AP. }\label{Table:tab1}
\begin{tabular}{p{12cm} >{\centering\arraybackslash}p{1.5cm} }
\hline
\multicolumn{2}{l}{\textbf{Models trained on baseline anomalies}}  \\
\hline
Baseline validation dataset ($N=200$) & 0.845  \\
Full validation dataset ($N=260$, baseline + smoothed edges) & 0.740  \\
\hline
\multicolumn{2}{l}{\textbf{Models trained on hard and smoothed edges}}  \\
\hline
Full validation dataset ($N=260$, baseline + smoothed edges) & 0.681 \\
\end{tabular}
\label{tab:first ablation}
\end{table}

Next, we attempted to assess whether training on complex anomaly shapes improved the performance in the validation set. We evaluated a new model trained with diverse shapes: cuboids, spheres and random shapes, chosen randomly. We denote this method as \textit{complex shapes} in Table \ref{Table:tab2}. Our experiments showed that using complex shapes improved AP in our validation set substantially, from $0.740$ to $0.817$ for models trained using anomalies with hard edges and from $0.681$ to $0.721$ on models trained using anomalies with a combination of hard and smoothed edges. Furthermore we found that ensembling two models, one trained with only hard edges and one with hard and smoothed-edge anomalies improved AP further to $0.881$.

\begin{table}[h!]
\centering
\caption{Ablations analyses on training using complex shapes (evaluated on the full abd. validation set). Metric is AP.}\label{Table:tab2}
\begin{tabular}{p{4.5cm} >{\centering\arraybackslash}p{2.5cm} >{\centering\arraybackslash}p{2.2cm} >{\centering\arraybackslash}p{2.2cm} }
\hline
 \textbf{Training set anomalies} & \textbf{Hard edges} & \textbf{+Smoothed edges} & \textbf{Ensemble}\\
\hline
Cuboids only & 0.740 & 0.681 & n/a  \\
Complex shapes & 0.817 & 0.721 & 0.881 \\
\hline
\end{tabular}
\end{table}

\begin{figure}[h!]
\centering
\includegraphics[width=1.\textwidth]{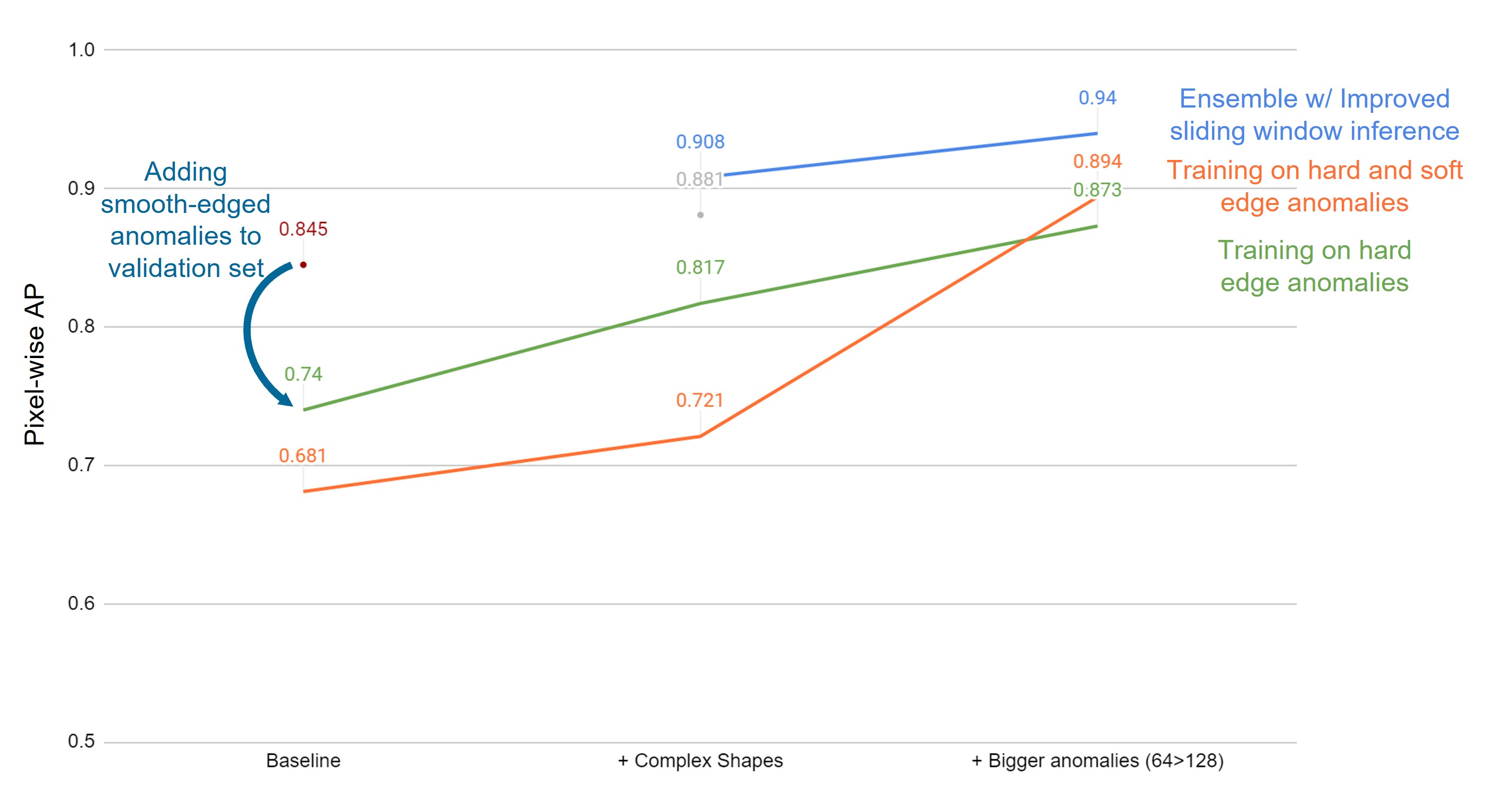}
\caption{Summary of individual proposed contributions and their impact on AP (abdominal validation set).} \label{fig6}
\end{figure}

Working towards our MOOD 2022 challenge submission, qualitative analysis showed that our method was under-performing in the bigger synthetic anomalies. Consequently, we doubled the anomaly size to $W_{fp} = H_{fp} = D_{fp} = 128$. We also increased the patch overlap in the sliding window inference from $0.25$ to $0.5$. Our final submission consisted of an ensemble of models trained on both hard- and smoothed-edge anomalies. These models achieved AP of $0.874, 0.894, 0.940$, for hard, smoothed and ensembled models respectively. In both abdominal and brain datasets we submitted an ensemble of 10 models, trained on hard- and smoothed-edge anomalies using 5 random data folds. Figure \ref{fig6} summarizes AP values for the different experiments evaluated.

\begin{figure}[h!]
\centering
\includegraphics[width=0.8\textwidth]{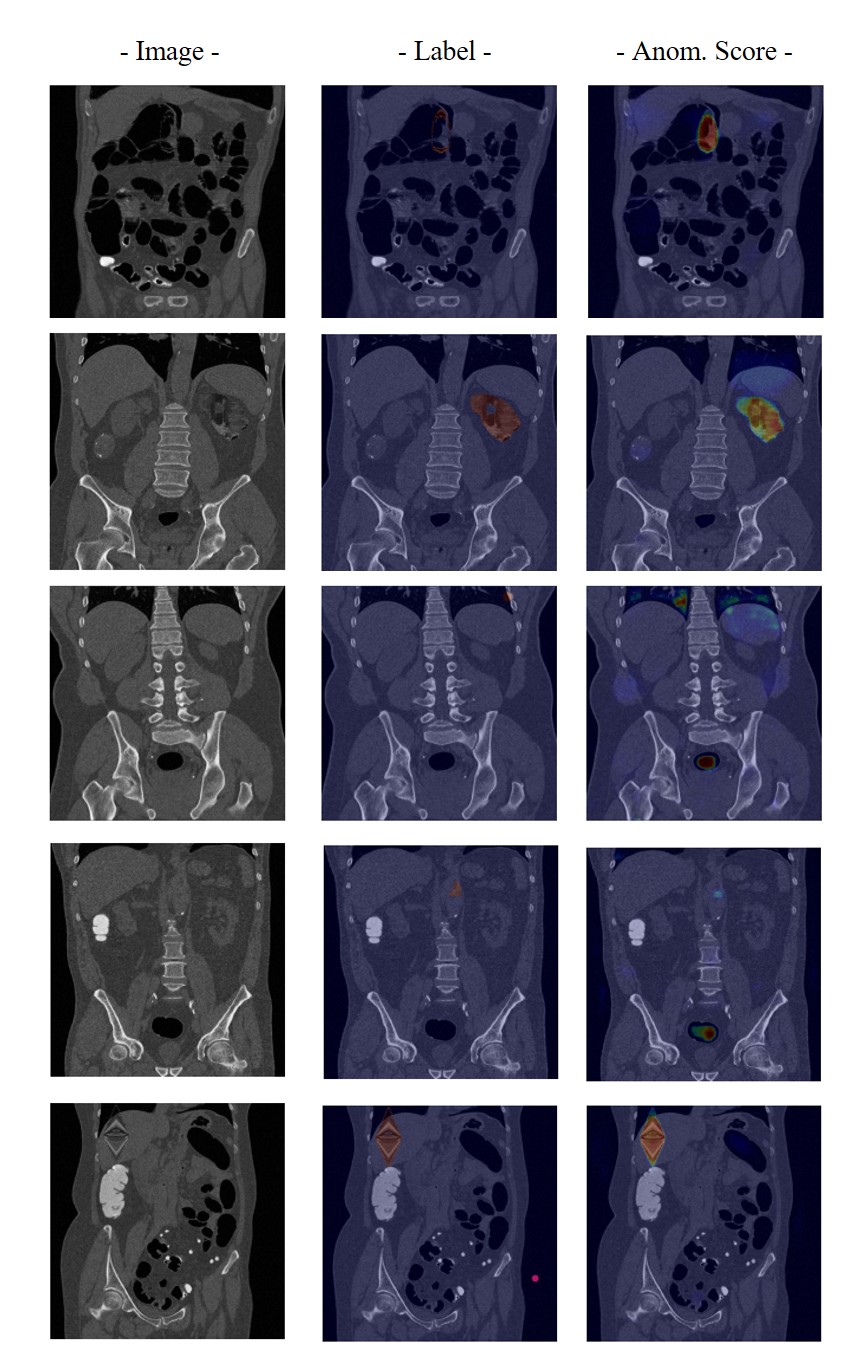}
\caption{Qualitative results from the MOOD 2022 challenge test set (abdominal CT). Columns show respectively test images, ground-truth label for the anomalies and pixel-wise anomaly score obtained by our method.} \label{fig7}
\end{figure}

\begin{figure}[h!]
\centering
\includegraphics[width=0.8\textwidth]{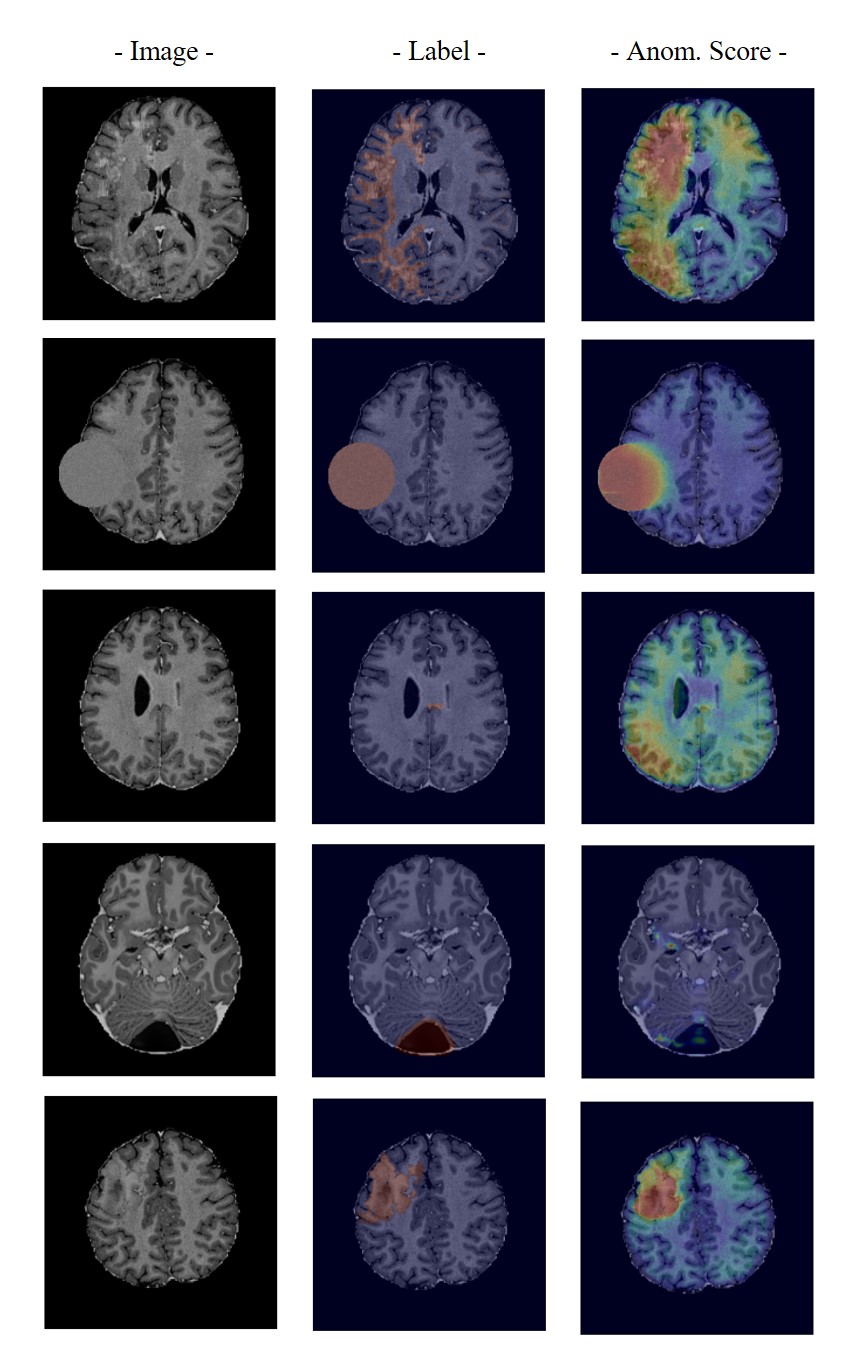}
\caption{Qualitative results from the MOOD 2022 challenge test set (brain MRI). Columns show respectively test images, ground-truth label for the anomalies and pixel-wise anomaly score obtained by our method.} \label{fig8}
\end{figure}

Our team submission achieved 1st place in both abdominal and brain datasets for the sample-wise task. We also achieved 1st and 4th place in abdominal and brain pixel-wise tasks respectively, which resulted as well in a 1st place overall. Qualitative examples received from the challenge organisers are included in Figures \ref{fig7} and \ref{fig8}. 

Our proposed method has showcased its effectiveness in the MOOD challenge, achieving a first position in both sample-wise and pixel-wise tasks. Despite the MOOD challenge results, it is not possible to compare quantitatively our method's performance with other state-of-the-art unsupervised anomaly detection methods since our experiments do not include quantitative comparisons. When comparing with other state-of-the-art methods, our simple method can generate plausible 3D synthetic anomalies more efficiently. Methods relying on Poisson image blending, such as PII \cite{Tan2021} and NSA \cite{Schluter2022}, also avoid gradients in the interpolation edges but are comparatively computationally expensive and more complex to implement. We leave for future work performing a comprehensive comparison against existing methods in clinically relevant settings where the potential enhancements that can be identified and incorporated.

It is important to note the inconsistent performance of our method across the different datasets of the pixel-wise task in the MOOD 2022 challenge. Specifically, in the pixel-wise task, our method achieved first position in the abdominal CT dataset while it \textit{only} achieved fourth position in the brain MRI dataset. The difference of performance might have been associated with the method's hyper-parameters which were fine-tuned only for the abdominal CT dataset and may not be optimal for the Brain MRI modality. The differences between abdominal CT and brain MRI modalities and anatomical structures hints at the necessity for customized hyper-parameter configurations. In future works we aim to explore how hyper-parameters influence performance across multiple image modalities. Automated hyper-parameter search techniques could be evaluated to when requiring to adapt our method to new and different modalities.

Finally, one of the inherent limitations with synthetic anomalies in a self-supervised learning (SSL) framework, such as ours, is the limited interpretability. While our method and similar SSL approaches efficiently detect and localize anomalies, they are less interpretable than other unsupervised anomaly detection methods, such as reconstruction or \textit{restoration} based methods. The interpretability of results is crucial in medical applications, where clinicians require not just the detection but an understanding of the underlying reasons to make informed decisions. Future works should seek to develop interpretability aspects of unsupervised anomaly detection methods based on synthetic anomalies.
  
\section{Conclusion} \label{Conclusions}
In this study, we built upon existing OoD detection methods based on synthetic anomalies introducing simple yet highly-effective improvements in the anomaly generation process, including using random shapes instead of cuboids and smoothing the interpolation edge of anomalies. Results in a synthetic validation set of abdominal CT scans showed a substantially improved performance over baselines, which was confirmed by the first position in both sample-and pixel-wise tasks in the MICCAI MOOD 2022 Challenge.

In the future, OoD detection could support radiological investigations to identify and localise anomalies and imaging artefacts. Effectively, these tools could work alongside the radiographers, offering a real-time and cost-effective additional assessment of the images, potentially reducing inattentional blindness \cite{DrewT2013} and increasing the overall detection performance.

\section{Disclosures} \label{Disclosures}
All authors declare that they have no conflicts of interest.

\section{Code, Data, and Materials Availability} \label{Code, Data, and Materials Availability}
The archived version of the code described in this manuscript can be freely accessed through the following GitHub repository: \href{https://github.com/snavalm/mood22}{\texttt{https://github.com/snavalm/mood22}}

\noindent 
The data utilized in this study were obtained from the Medical Out-of-Distribution Challenge (MOOD) \cite{MOOD2020}. Data can be obtained upon registration in the MOOD Challenge portal \href{https://www.synapse.org/#!Synapse:syn21343101/wiki/599515}{\texttt{MOOD Challenge submission portal}}

\newpage
%
%

%
%
%
\bibliographystyle{splncs04}
\bibliography{ref.bib}
\end{document}